\documentclass[letterpaper, 10 pt, conference]{ieeeconf}  

\IEEEoverridecommandlockouts                              

\usepackage{graphics} 
\usepackage{epsfig} 
\usepackage{subcaption}
\usepackage{amsmath} 
\usepackage{amssymb}  
\usepackage{verbatim}
\usepackage{duckuments}
\usepackage{adjustbox}
\usepackage{wrapfig}
\usepackage{xcolor}
\usepackage{array}

\title{\LARGE \bf
 Immersive Teleoperation Framework for Locomanipulation Tasks
}

\author{Takuya Boehringer, Jonathan Embley-Riches, Karim Hammoud, Valerio Modugno, Dimitrios Kanoulas
\thanks{All the authors are with the Department of Computer Science, University College London, WC1E 6BT, London, UK.{\tt\small \{takuya.boehringer.21,jonathan.embley- riches.22,karim.hammoud.23, v.modugno,d.kanoulas\}@ucl.ac.uk}}%
}

\begin{document}

\maketitle
\thispagestyle{empty}
\pagestyle{empty}

\begin{abstract}

Recent advancements in robotic loco-manipulation have leveraged Virtual Reality (VR) to enhance the precision and immersiveness of teleoperation systems, significantly outperforming traditional methods reliant on 2D camera feeds and joystick controls. Despite these advancements, challenges remain, particularly concerning user experience across different setups. This paper introduces a novel VR-based teleoperation framework designed for a robotic manipulator integrated onto a mobile platform. Central to our approach is the application of Gaussian splattering, a technique that abstracts the manipulable scene into a VR environment, thereby enabling more intuitive and immersive interactions. Users can navigate and manipulate within the virtual scene as if interacting with a real robot, enhancing both the engagement and efficacy of teleoperation tasks. An extensive user study validates our approach, demonstrating significant usability and efficiency improvements. Two-thirds (66\%) of participants completed tasks faster, achieving an average time reduction of 43\%. Additionally, 93\% preferred the Gaussian Splat interface overall, with unanimous (100\%) recommendations for future use, highlighting improvements in precision, responsiveness, and situational awareness. Finally, we demonstrate the effectiveness of our framework through real-world experiments in two distinct application scenarios, showcasing the practical capabilities and versatility of the Splat-based VR interface.

\end{abstract}

\section{INTRODUCTION}






Advancements in robotic teleoperation systems have been central in the expansion of human capabilities into remote or hazardous environments. Traditional teleoperation interfaces often rely solely on joystick-based controls and visual feedback from cameras, each with inherent limitations that can reduce precision and immersion~\cite{Kourosh2023}. Despite significant advancements in the applications of Virtual Reality (VR) and Augmented Reality (AR) in the robotic teleoperation context~\cite{penco2024mixed, solanes2022virtual, galarza2023virtual}, there remains substantial untapped potential to enhance these systems further. Moreover, there are several frameworks that aim to address aspects of robotic loco-manipulation tasks, defined as the capability to move within an environment and dynamically interact with objects present in the scene~\cite{penco2019multimode}. Yet, these frameworks often lack comprehensive integration with advanced VR technologies that can adapt across various hardware platforms and operational environments. To bridge this gap, we introduce a comprehensive VR-based framework specifically designed for loco-manipulation tasks.

In this paper, we introduce a novel teleoperation framework that enhances task precision and effectiveness through the integration of advanced VR with Gaussian splattering, as introduced in~\cite{kerbl20233d}. Unlike traditional VR interfaces that rely on visual feedback~\cite{penco2024mixed}, our approach utilizes (but is not limited to it) Gaussian splattering to create a high-fidelity, three-dimensional representation of remote environments, specifically designed to manage occlusions effectively. This rendering technique not only provides deeper spatial coherence but also enables operators to freely adjust their viewpoint within the VR environment, significantly aiding in tasks that require precision manipulation and reaching around occlusions.

\begin{figure}[t] 
  \centering
  \includegraphics[width=0.45\textwidth]{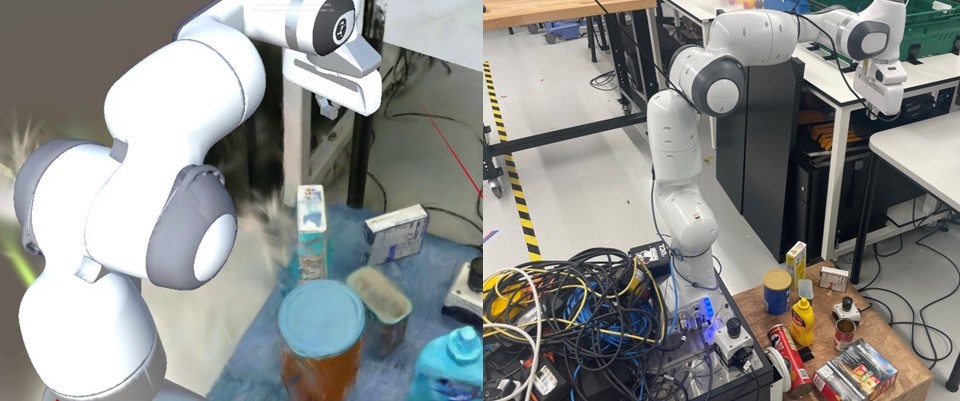} 
  \caption{In this picture, we showcase an example of how our proposed framework is utilized. On the left, there is a virtual reality representation of the robot and the scene, reconstructed using Gaussian splattering. On the right, the actual robot, mounted on a mobile base, is depicted in the real environment during the manipulation phase}
  \label{fig:frontispiece}
\end{figure}

Our framework operates within a dual-phase strategy: initial navigation using a conventional control interface with visual feedback for locomotion, followed by detailed manipulation tasks within the VR environment, as shown in Figure~\ref{fig:frontispiece}. Here, operators can directly manipulate a virtual representation of the robot's arm, allowing for intuitive control as they simply drag the model within the VR space. This method enhances the operator's sense of presence and engagement and improves the precision of remote manipulations. 
These results were quantified in a user study, testing the effectiveness and intuitiveness of the VR interface. The interface was compared to a simple baseline, demonstrating the advantages of our system via a pick-and-place task in a complex, visually obstructed scenario.


The contributions of this work can be summarized as follows:
\begin{itemize}
    \item \textit{Two-Stage Framework for Loco-manipulation with VR interaface}:
    We present a two-stage framework for loco-manipulation that integrates seamlessly with a VR interface. The initial stage uses a conventional control interface for navigation, followed by detailed manipulation tasks within the VR environment. This setup enables intuitive control through direct manipulation of a virtual robot's arm, enhancing usability.

    \item \textit{Enhanced Immersion and Precision Manipulation}: Our framework enhances immersion and precision manipulation by integrating advanced VR with Gaussian splattering techniques. This combination produces a high-fidelity, three-dimensional representation of remote environments, effectively managing occlusions and enhancing spatial coherence. It significantly enhances task performance, especially for precise control. 

    \item \textit{User Study}:
    This validates the effectiveness of our novel interface, quantifying the enhanced task performance and thereby demonstrating the benefits of integrating Gaussian Splatting and VR for teleoperation.
\end{itemize}

The rest of this paper is organized as follows: 
Section \ref{sec:RW} reviews the relevant literature and existing frameworks for teleoperation and VR-based systems. 
Section \ref{sec:framework} presents our proposed framework for immersive teleoperation, detailing the integration of VR with Gaussian splattering and the two-stage approach for locomanipulation tasks.
Section \ref{sec:results} describes the experimental results, showcasing the performance and benefits of our system quantitatively in our user study scenarios blue and qualitatively in various teleoperation. 
Finally, Section \ref{sec:conclusion} concludes the paper and outlines potential future work directions.

\section{RELATED WORK}\label{sec:RW}

In recent years, numerous papers have been published proposing novel teleoperation frameworks. In \cite{yang2024ace}, the authors introduce a system designed for precise control of robotic hands and grippers through a physical, exoskeleton-based interface. It employs 3D-printed exoskeletons and vision-based camera systems to capture accurate hand poses, which are then transferred to various robotic platforms. In \cite{cheng2024open}, the authors describe Open-TeleVision, a teleoperation system that emphasizes immersive and active visual feedback using a stereoscopic camera system mounted on the robot's head, which mirrors human head movements. This design allows the operator to see from the robot's first-person perspective, enabling them to actively explore the robot’s workspace and engage in tasks requiring precise manipulation and depth perception. The system utilizes inverse kinematics to directly mirror the operator's hand and arm movements, providing an intuitive control experience that simulates embodying the robot. Similarly, in \cite{qin2023anyteleop}, AnyTeleop is introduced. According to the authors, it is designed as a general, vision-based teleoperation system that uses low-cost camera sensors to support a wide range of camera configurations like RGB and depth cameras and is capable of handling multiple cameras. This system is particularly noted for its broad generalization capabilities across various robot models, both in simulated and real environments, making it highly adaptable to different hardware setups. All these papers solely focus on the manipulation aspect of robotic teleoperation, and the use of VR is not explored to enhance the user experience. In \cite{penco2019multimode}, the authors propose an innovative teleoperation framework for humanoid robots, designed to tackle the challenges of operating in hazardous environments. Although the method is designed to explicitly tackle complex loco-manipulation tasks, it lacks integration with a VR interface.

Numerous frameworks have been proposed that integrate teleoperation systems with advanced visualization technologies. VR has been instrumental in enhancing user immersion and control, which is crucial for improving teleoperation performance. In \cite{penco2024mixed}, the authors present a teleoperation framework that integrates mixed reality and assistive autonomy to improve the control of humanoid robots. This system employs Probabilistic Movement Primitives, Affordance Templates, and object detection technologies to enhance the efficiency and effectiveness of humanoid robot operations, particularly in tasks requiring intricate object manipulation and environment interaction. The paper presents an attempt to tackle the same problem from a different perspective, focusing more on the problem of preemptive motion generation. No mention is made regarding locomanipulation tasks, and the method requires continuous visual feedback to function. In \cite{naceri2021vicarios}, the authors introduce a VR-based interface designed to enhance intuitive and real-time control in remote robotic teleoperation, particularly in hazardous environments such as nuclear decommissioning, disaster response, and surgery. This system aims to overcome limitations associated with traditional teleoperation interfaces by providing a more immersive and interactive environment. Unlike this method, our framework leverages more photorealistic and dynamic visualization, which can provide a deeper level of detail and realism to make tasks such as precise manipulation more feasible to perform. In \cite{stotko2019vr}, the authors design a novel virtual reality-based system to enhance remote robotic teleoperation. This VR system addresses the limitations of traditional video-based teleoperation, such as limited immersion and situational awareness, by leveraging VR technology to create an interactive virtual representation of the environment. It utilizes RGB-D data, processed using SLAM (Simultaneous Localization And Mapping) techniques, to construct a real-time, comprehensive 3D model of the environment, which is then streamed to VR headsets. The method proposed in \cite{stotko2019vr} primarily focuses on augmenting awareness for navigating in unknown environments rather than concentrating on locomanipulation, which is the main focus of our method. The work presented in \cite{nakanishi2020towards} explores the development of a teleoperation system that uses virtual reality (VR) to control Toyota's Human Support Robot (HSR). The system aims to provide a more natural and immersive method for human operators to control the HSR, particularly for assistance in home settings. It integrates a head-mounted VR device with stereo vision from the robot’s cameras and uses a singularity-robust inverse kinematics algorithm to enhance movement coordination between the robot's base and arm. Unlike the aforementioned framework, our work focuses more on the problem of fine manipulation with occlusion, while providing navigation capability to the operator. In \cite{solanes2022virtual}, the authors introduce a minimalistic VR interface designed to simplify human-robot interaction by allowing the robot to autonomously handle collision avoidance while the human operator focuses on guiding the robot in complex environments. Our work solely relies on a photorealistic VR interface with an intuitive teleoperation modality, which is very different from the simplistic interface proposed in this work. In \cite{galarza2023virtual}, the authors discuss a VR-based system designed to enhance the teleoperation of the KUKA youBot robot using HTC Vive Pro 2 goggles. Although the proposed system employs a Unity interface for the VR headset, the proposed visualization is minimal and is only employed for a fixed robot. Both in \cite{de2021leveraging} and \cite{kuo2021development}, the authors discuss the development and effectiveness of Virtual Reality frameworks aimed at improving robotic teleoperation, especially for tasks demanding precise control. These frameworks capture environmental and robot data using RGB-D cameras, which are then reconstructed in 3D and relayed to a remote operator through a VR headset. These works showcase how the use of immersive VR interfaces can effectively enhance the user's capability to perform teleoperation and present several elements of contact with our framework.

\begin{figure}[t] 
  \includegraphics[width=0.45\textwidth]{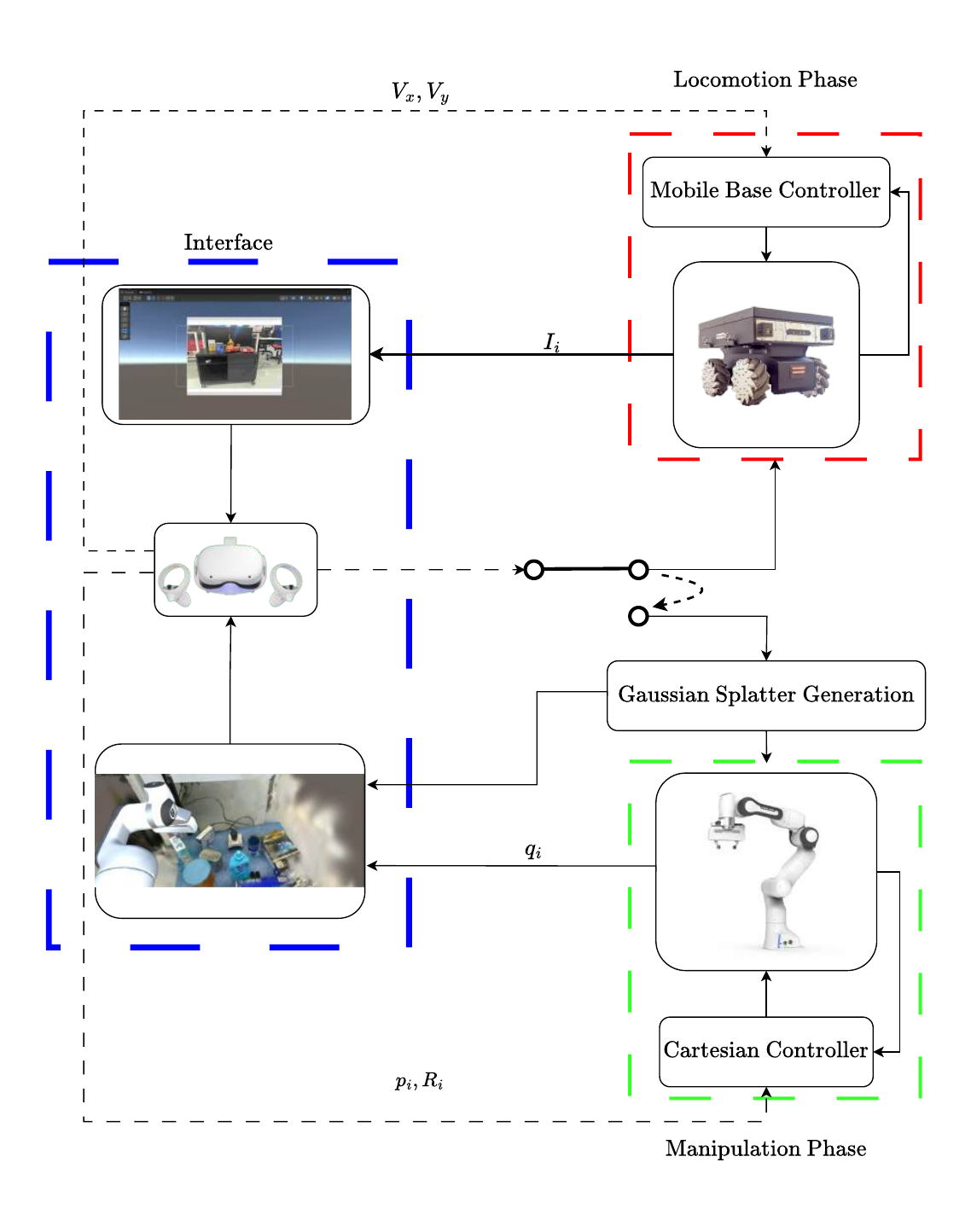}
  \caption{This image details the structural scheme of the teleoperation framework} 
  \label{fig:framework_scheme}
\end{figure}

Finally, several authors have explored the use of photorealistic representations such as Neural Radiance Fields (NeRFs) and Gaussian Splatting \cite{patil2024radiance}, \cite{lu2024manigaussian}. In \cite{patil2024radiance}, the authors investigate the integration of both NeRFs and Gaussian Splatting into robotic teleoperation systems to enhance visualization capabilities. Unlike this work, we not only explore the advantages of VR photorealistic representation but also examine how this can effectively enhance fine manipulation capability in unstructured scenarios. The method introduced in \cite{lu2024manigaussian} effectively extends the classic Gaussian Splatting approach for dynamic environments, which, while not directly used for teleoperation, showcases the potential development of this technology for dynamic settings.

\section{FRAMEWORK FOR IMMERSIVE TELEOPERATION}\label{sec:framework}

This section presents the design and operation of the high-level teleoperation framework. We first provide an overview of the system architecture, followed by a detailed explanation of each stage, illustrating how the sub-modules interact and communicate within the system.

The framework operates in two distinct modes: the \textit{Locomotion Phase} and the \textit{Manipulation Phase} (see Figure~\ref{fig:framework_scheme}). Initially, the system functions in the locomotion phase, where the human operator controls the robot's base using a teleoperation interface. In this mode, the operator utilizes a video feed from the robot’s base, combined with VR controllers, to navigate the robot toward the desired manipulation site. Once the robot reaches the target location, the operator initiates the manipulation phase, during which the robotic arm is teleoperated through the VR interface. To facilitate this, a Gaussian splatter is generated to create an exact copy of the operational scene in the VR environment, along with an aligned model of the manipulator. This provides the user with a fully immersive VR representation that facilitates the interaction of the user with the real world providing an intuitive interface.

\subsection{Virtual Reality Interface}
The human operator interacts with the system via the VR interface. The primary control mechanism consists of joystick inputs, used to switch between operation phases and to command both the base and the manipulator. Phase switching is deliberately controlled by physical inputs on the joystick, rather than within the VR environment, to ensure that user actions within the virtual space do not inadvertently interfere with the system's phase transitions.

The Gaussian splat is implemented as a visual effect within the Unity scene. Since the Gaussian splatter is a neural rendering of a point cloud, it is naturally suited to implementation via a node-based particle system such as Unity's VFX graph.

The Unity engine communicates with the robot’s base and manipulator through the Robot Operating System (ROS). The robot’s onboard computer hosts a TCP endpoint, through which all ROS communications from Unity are routed. During the locomotion phase, joystick commands are published over ROS, while the video stream from the robot’s base is subscribed to in real-time. In the manipulation phase, a feedback loop is employed to control the manipulator via the VR interface. When the operator moves the VR representation of the end effector, the corresponding Cartesian coordinates are transmitted to the manipulator controller over ROS, commanding the real-world manipulator to reach the new position. Simultaneously, the updated joint angles are transmitted back to Unity, allowing it to adjust the VR model of the manipulator to mirror the physical robot's movements.

This interface is highly adaptable to various robotic platforms. The Gaussian splatter process remains unaffected by the specific manipulator used for data collection. Adapting the system to new bases or manipulators typically requires only minor adjustments, such as modifying ROS topic names in Unity and the robot's onboard system, and importing a new manipulator model into the Unity scene. Most commonly used robotic manipulators have pre-existing URDF models, and the number of joint angles to be adjusted during teleoperation can be easily configured.

\subsection{Locomotion Phase}
The starting state of the framework is in the locomotion phase. The base sends an image stream to the interface and sends positional commands to the base. In this way, the human operator controls the base movement while navigating via a 2D video stream. The 2D video is projected onto a screen at a fixed position in VR to minimize operator motion sickness due to the reduced Field Of View of the operator~\cite{Chang2020}. The base itself uses a controller to ensure steady movement. For generality, regardless of the mobile base type, we assume that the mobile base can be commanded by providing a Cartesian velocity command. The controller receives simple velocity commands from the interface in the form of $(V_x, V_y)$.

\subsection{Gaussian Splatter for Environment Reconstruction}
\begin{figure}[t] 
  \centering
  \includegraphics[width=0.45\textwidth]{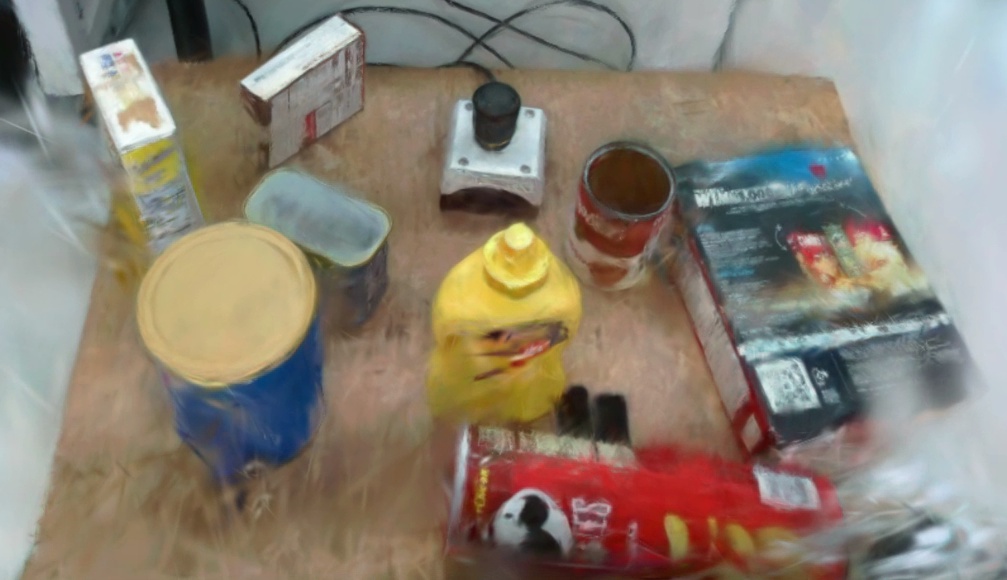} 
  \caption{In this image, we present an example of a scene reconstructed in virtual reality using Gaussian splatting techniques}
  \label{fig:splat example}
\end{figure}
In order to switch the framework to the manipulation phase, the framework must first generate a Gaussian splatter to be placed in the VR environment, such as that in Figure~\ref{fig:splat example}. This is itself completed in three stages. First, the manipulator moves to a series of pre-planned positions to obtain overlapping images of the scene. Next, these images are used as input for Structure-from-Motion (SfM) to detect features, estimate camera poses, and use triangulation to calculate 3D points, followed by bundle adjustment for refinement~\cite{snavely2006photo}. Finally, the Gaussian splatter itself is trained, using neural rendering to generate a 3D model that can be used for VR teleoperation. The scene is represented as a set of Gaussian distributions, each with a position, size, orientation, and color. The Gaussians are parameterized by their mean (position in 3D space), covariance matrix (which defines their shape and spread), and attributes like color and opacity. During rendering, each Gaussian is projected onto the 2D image plane as an elliptical ``splat''. These splats are combined to form the final rendered image \cite{kerbl20233d}.

\subsection{Manipulation Phase} 
Once the Gaussian splat is generated, the system transitions into the manipulation phase. During this phase, a feedback loop is initiated wherein the human operator controls the movement of the manipulator's end-effector within the VR environment. Several teleoperation control strategies were considered during the development process. Many existing systems link controllers to a frame of reference anchored in or near the operator's first-person view of the scene. However, this approach limits the operator's ability to fully leverage the immersive capabilities of the VR environment, as it constrains both movement and field of view. In contrast, our framework adopts a game-like interface where the 3D reconstruction of the robotic arm and the operational scene are positioned directly in front of the operator. The operator can navigate freely around this 3D reconstruction within the VR environment and interact with it by `grabbing' the end-effector and `dragging' it to move the real-world manipulator. 
The camera views from the base and the end-effector are overlayed on the VR view so the user can still see the changes to the real-world scene during teleoperation through the static 3D reconstruction.

\begin{figure*}[] 
  \includegraphics[width=\textwidth,height=\textheight,keepaspectratio]{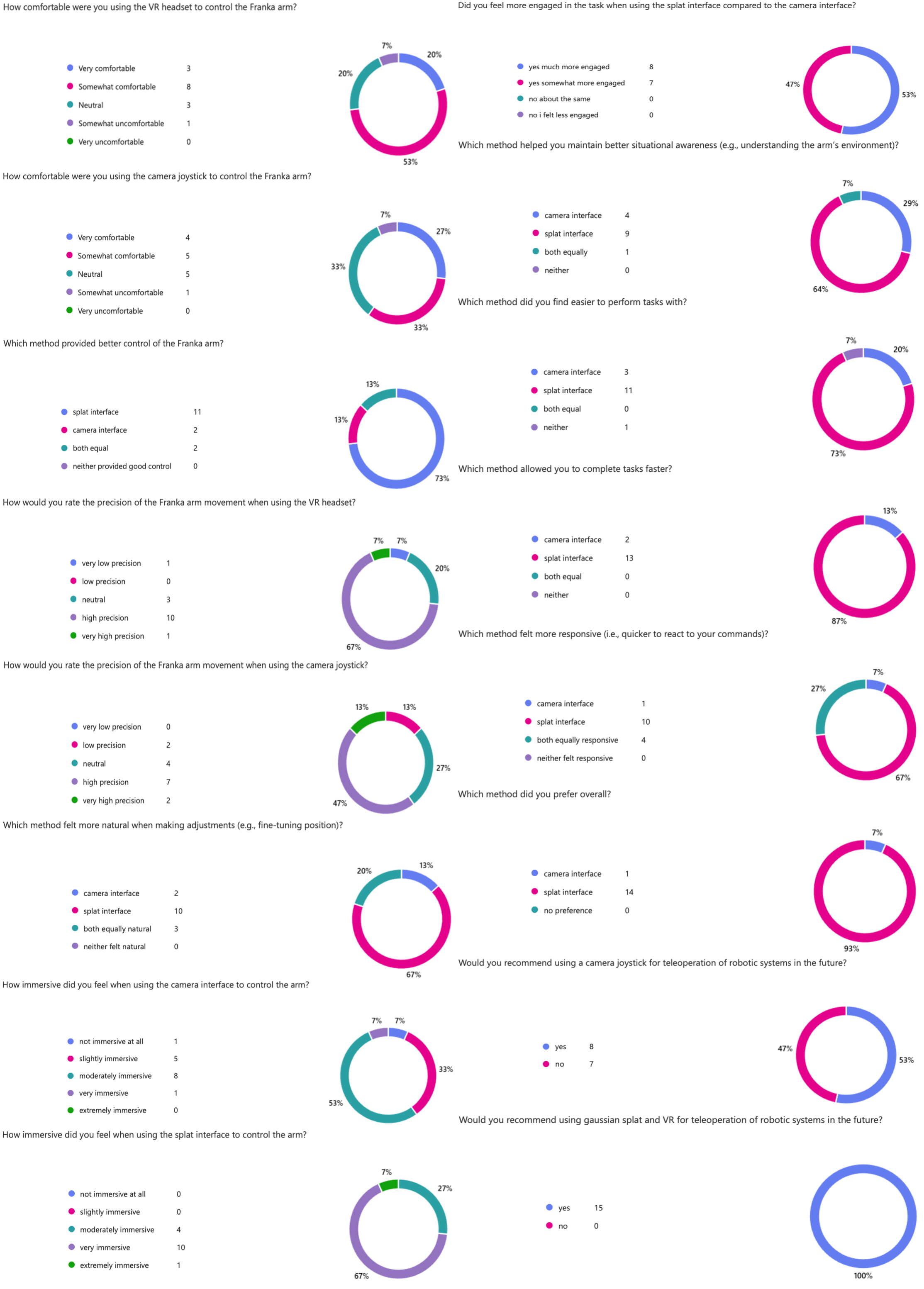} 
  \caption{Results from the non-scaled questions in the user study.}
  \label{fig:user_study_nonscaled_questions}
\end{figure*}
\section{RESULTS AND DISCUSSION}\label{sec:results}
The framework was rigorously tested in a user study and compared to a baseline that relied only on the feed from two cameras. The results demonstrated that the proposed framework outperformed the baseline. Additionally, we conducted evaluations on a real robot across two tasks designed to assess its functionality in both operational phases.
\subsection{User Study Design and Simulation Environment}
The user study involved participants performing a pick-and-place task in VR using two different interfaces. The object to be moved was in a cluttered scene, occluded and obstructed by other objects. This setup was an accurate representation of a use case of the full framework, with a Gaussian splat of the objects to be grabbed by the URDF manipulator model controlled by the user.

15 participants took part in the user study. Each participant was given as much time as needed to test out the controls in both interfaces before the trial started. Once they were ready, participants were asked to attempt the task using the baseline interface and then again using the splat interface. Finally, they were asked to fill out a questionnaire comparing the merits and drawbacks of the two interfaces.

The questionnaire consisted of asking if the participant had prior VR experience, the time taken for both interfaces (all trials were timed) and a series of questions about the experience. The first 16 questions were scaled, asked about each interface. These questions can be seen in figure \ref{fig:user study data}. Next, 16 non-scaled questions were asked, as shown in figure \ref{fig:user_study_nonscaled_questions}, including which interface was preferred overall and whether they would recommend either interface for future use.

\begin{figure}[h]
    \centering
    \includegraphics[width=0.45\textwidth]{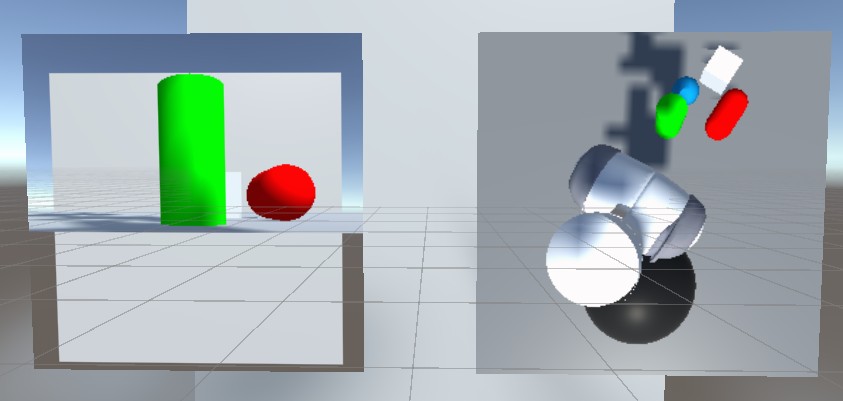}
    \caption{The baseline interface, displaying the camera view from the base of the manipulator on the left and the camera view from the end-effector on the right.}
    \label{fig:baseline_interface}
\end{figure}

The baseline interface, as shown in figure \ref{fig:baseline_interface} provided two camera views from the manipulator in the VR environment. One view was from a camera under the base of the manipulator and the other from a camera attached to the end-effector. The user teleoperated the manipulator using the joysticks on the VR controllers, mimicking a typical manipulator teleoperation interface.

\begin{figure}[h]
    \centering
    \includegraphics[width=0.45\textwidth]{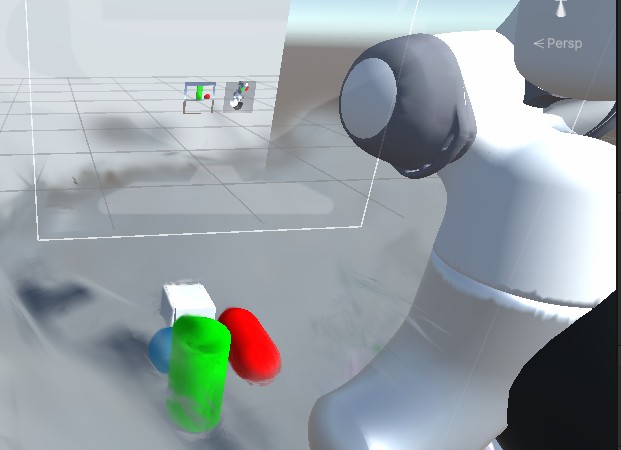}
    \caption{The splat interface, showing the two camera views overlaid on the display. The user, physically present in the VR environment, sees the static splat overlaying the objects rather than the actual objects themselves and can manipulate the manipulator End Effector shown in VR}
    \label{fig:splat_interface}
\end{figure}

The splat interface, as shown in figure \ref{fig:splat_interface}, instead placed the user in the VR scene with the Gaussian splat and the manipulator, allowing the user to move and peer around the scene. This also provided the user with the option to control the manipulator by 'grabbing' the end-effector and 'dragging' it with their VR controller. The two camera views from the baseline interface were still provided, overlayed on the VR display. Since the splat is static, the user can't see when an object has been grabbed since the main VR view shows the splat and not the real objects. The two camera views show the real objects, acting as references to the user during the picking and placing.


\subsection{Quantitative Results}
The findings from the user study strongly support the effectiveness and intuitiveness of the Gaussian Splat-based VR interface, demonstrating statistically and practically significant advantages over the traditional camera joystick system across usability, efficiency, and user engagement metrics. Notably, two-thirds (66\%) of participants completed their initial task significantly faster with the Splat interface, achieving an average time reduction of 43\% compared to the joystick baseline. This advantage was particularly pronounced among experienced VR users, with 9 out of 10 showing faster performance, suggesting the interface effectively leverages users' existing spatial and interaction skills. Quantitative results reinforced these findings, with participants rating the Splat interface higher in ease of use (5.53 vs. 4.33), visual feedback clarity (5.80 vs. 4.87), situational awareness (6.13 vs. 4.13), immersion (5.87 vs. 4.33), and reduced cognitive load (5.60 vs. 4.40). Qualitative feedback further underscored the Splat system's superiority, with 93\% of participants expressing overall preference and 100\% recommending it for future use, compared to only 53\% recommending the joystick. Specifically, participants preferred the Splat interface over the joystick in terms of control precision (11 participants preferred Splat versus 2 for joystick), task completion speed (13 preferred Splat versus 2 for joystick), and responsiveness (10 preferred Splat versus 1 for joystick), particularly emphasizing enhanced situational awareness crucial for effective teleoperation.

\begin{center}
\begin{figure}[]
    \begin{tabular}{ | m{15em} | m{1cm}| m{1cm} | }  
    \hline
    Question: & Average Rating (Baseline): & Average Rating (Splat): \\ 
    \hline
    I found this control interface easy to use & 4.33 & 5.53 \\ 
    \hline
    The control inputs accurately translated to robot movements & 5.60 & 5.87 \\ 
    \hline
    I felt fully immersed in the environment while using this system & 4.20 & 5.88 \\ 
    \hline
    I had a good sense of the robot’s position and orientation at all times & 4.13 & 6.13 \\ 
    \hline
    The visual feedback provided a clear understanding of the task environment & 4.87 & 5.80 \\ 
    \hline
    I felt as if I was physically present in the robot’s operational space & 4.33 & 5.87 \\ 
    \hline
    Controlling the robot with this interface required minimal mental effort & 4.40 & 5.60 \\ 
    \hline
    I would be interested in using this system regularly for controlling robotic tasks & 4.47 & 6.13 \\ 
    \hline
    \end{tabular}
    \caption{
    Results from the scaled questions in the user study, evaluating the usability and intuitiveness of the two interfaces. A 7 point scale was used where 1 means strongly disagree and 7 means strongly agree.}
    \label{fig:user study data}
\end{figure}
\end{center}

\subsection{Experimental Setup}
Our full framework has also been effectively tested on a robotic stack composed of:
    \begin{itemize}
        \item \textit{Robotnik Summit-XL:} Provides the base for the robotic system. The built-in computer on the Summit-XL publishes ROS nodes for controlling the base and streaming the 2D camera feed.
        \item \textit{Franka Emika:} Attached to the Summit-XL and controlled through the onboard computer and Intel NUC.
        \item \textit{Intel NUC:} Acts as a central hub for controlling the manipulator and processing camera data. It has Ethernet connections to both the Summit-XL base and the manipulator and a USB connection to the Intel RealSense camera.
        \item \textit{Meta Quest 2:} This is the VR interface that we are using in our application. The device is connected to a computer that executes the Quest link and the Unity programs.  A dedicated TCP bridge connects the Unity software to the Nuc. The computer that runs the Unity programs is equipped with a 3090 NVidia GPU to handle the training of the Gaussian splatter. 
        \item \textit{Intel RealSense D435F Camera:} Mounted on the manipulator’s end-effector, it captures a series of images of the scene to be used for 3D reconstruction and to build the Gaussian Splatting.
    \end{itemize}






\begin{figure}[b] 
  \centering
  \includegraphics[width=0.45\textwidth, angle=180]{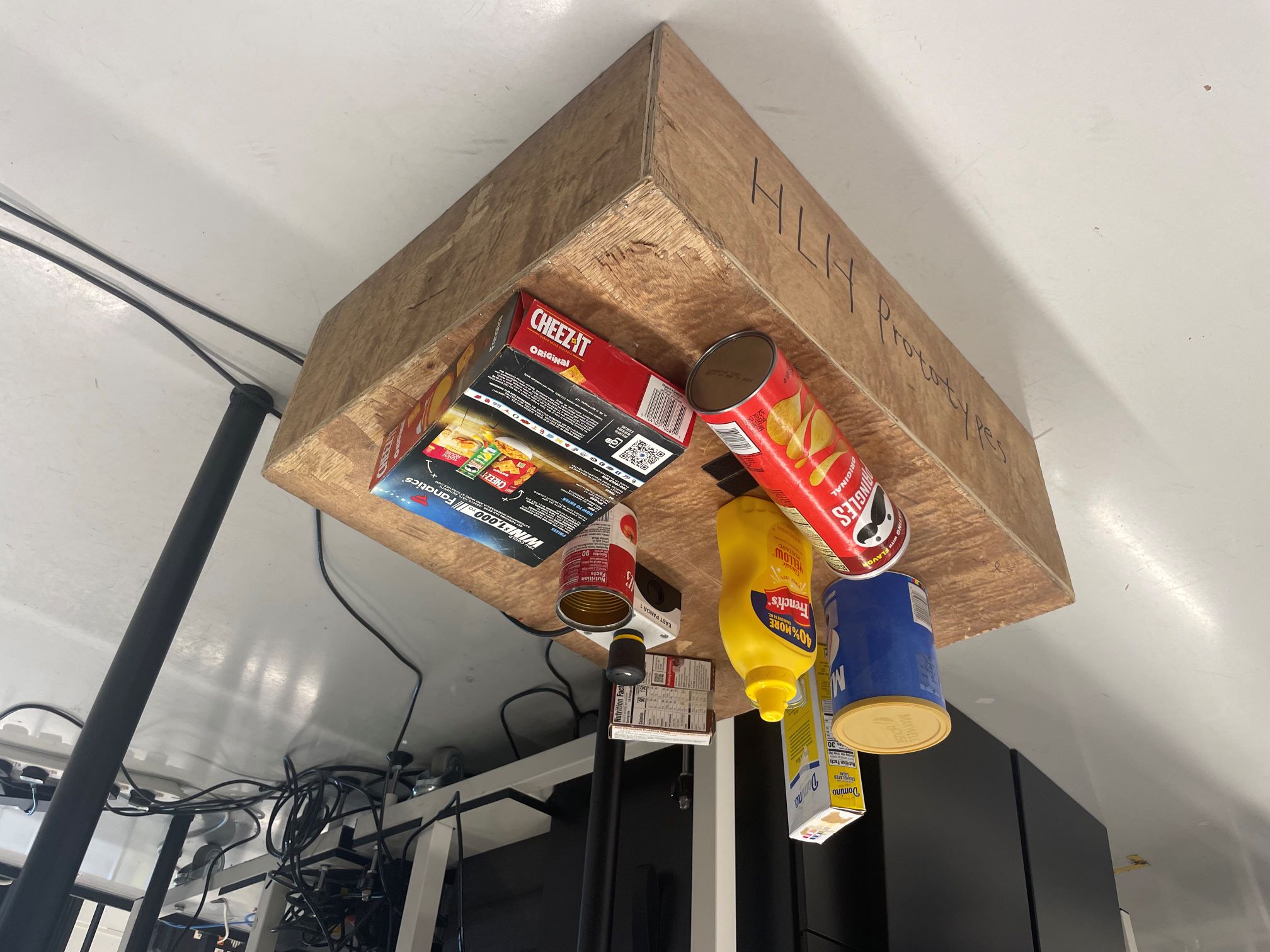} 
  \caption{In this image, we illustrate the setup of the first operational scenario, where the robot reaches a button in a cluttered environment. The button is placed on a base that obstructs the view from the robot's camera}
  \label{fig:sctrucut enario1}
\end{figure}


\begin{figure}[h] 
  \centering
  \begin{subfigure}{0.16\textwidth}
    \includegraphics[width=\linewidth, angle=180]{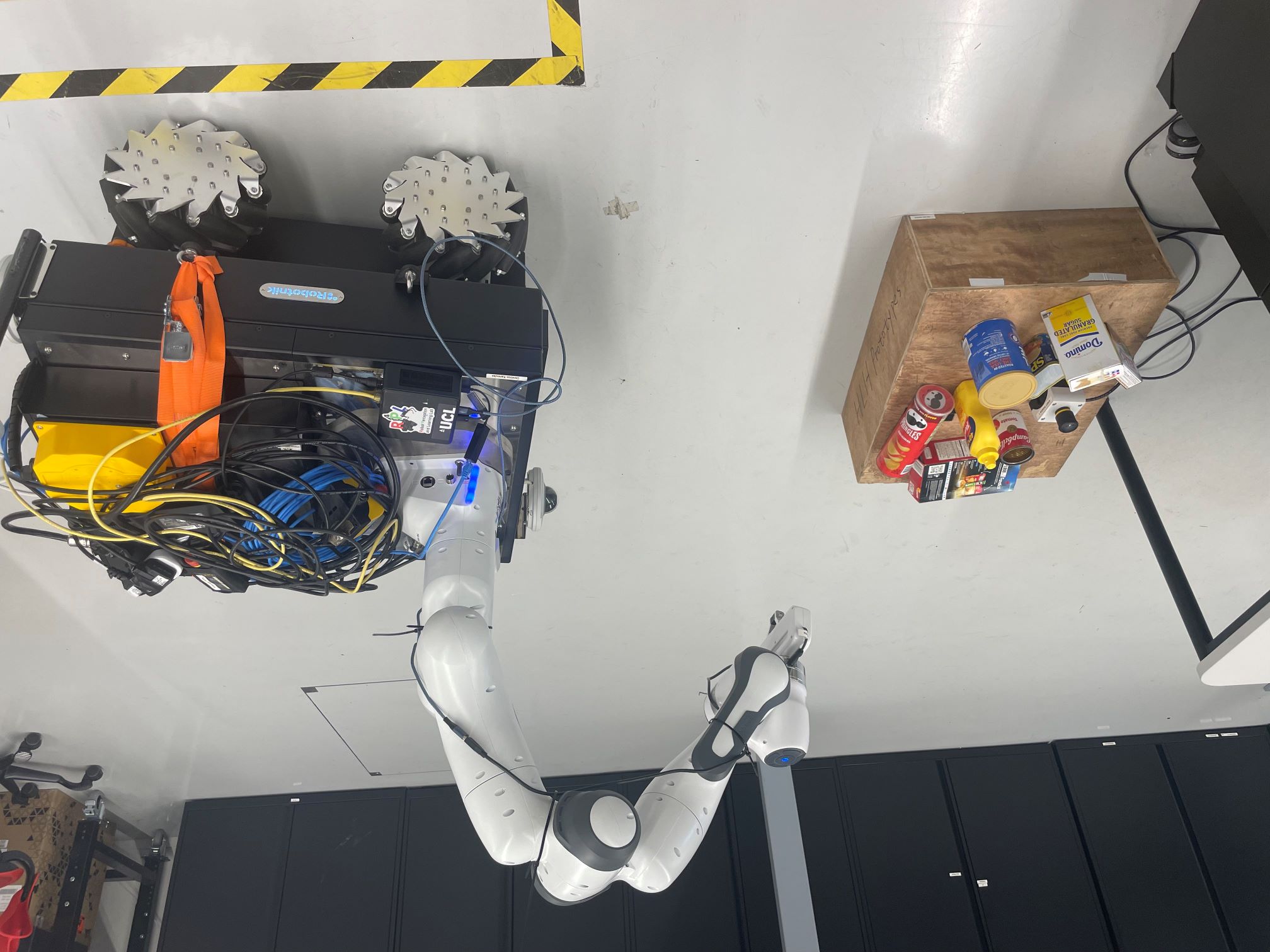} 
    \caption{}
    \label{fig:exp1sub1}
  \end{subfigure}%
  \begin{subfigure}{0.16\textwidth}
    \includegraphics[width=\linewidth, angle=180]{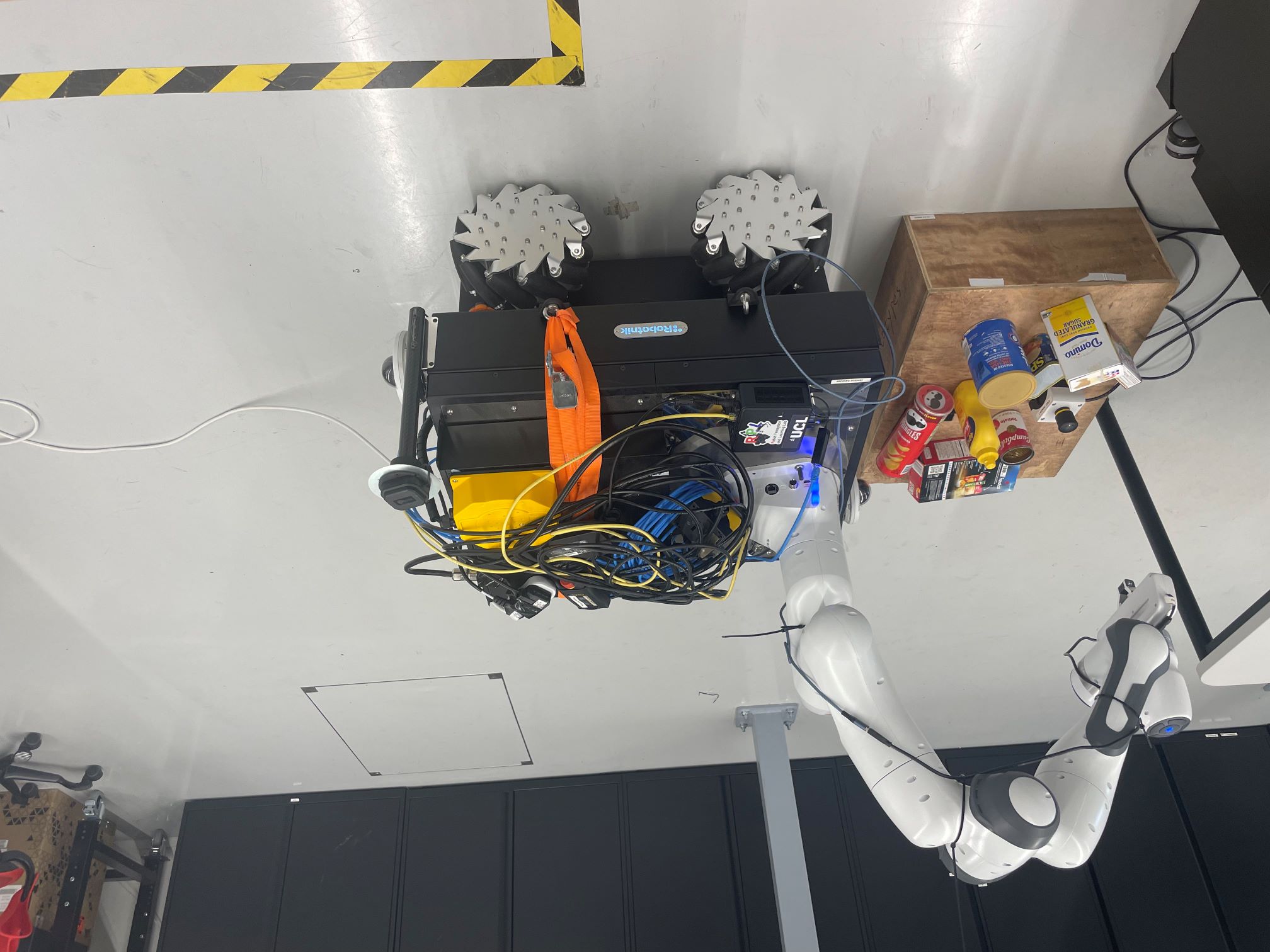} 
    \caption{}
    \label{fig:exp1sub2}
  \end{subfigure}%
  \begin{subfigure}{0.16\textwidth}
    \includegraphics[width=\linewidth, angle=180]{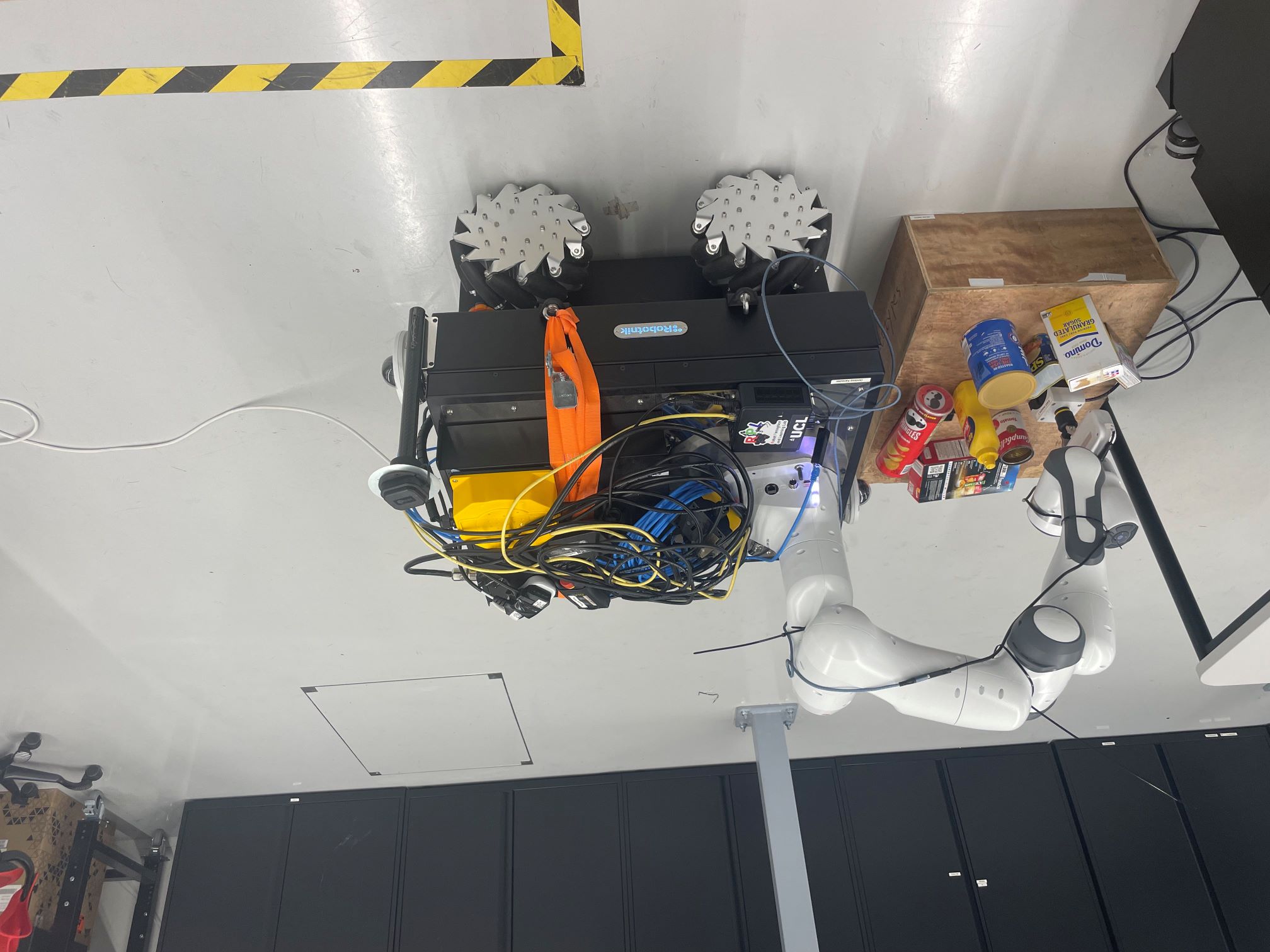} 
    \caption{}
    \label{fig:exp1sub3}
  \end{subfigure}
  \begin{subfigure}{0.16\textwidth}
    \includegraphics[width=\linewidth , angle=180]{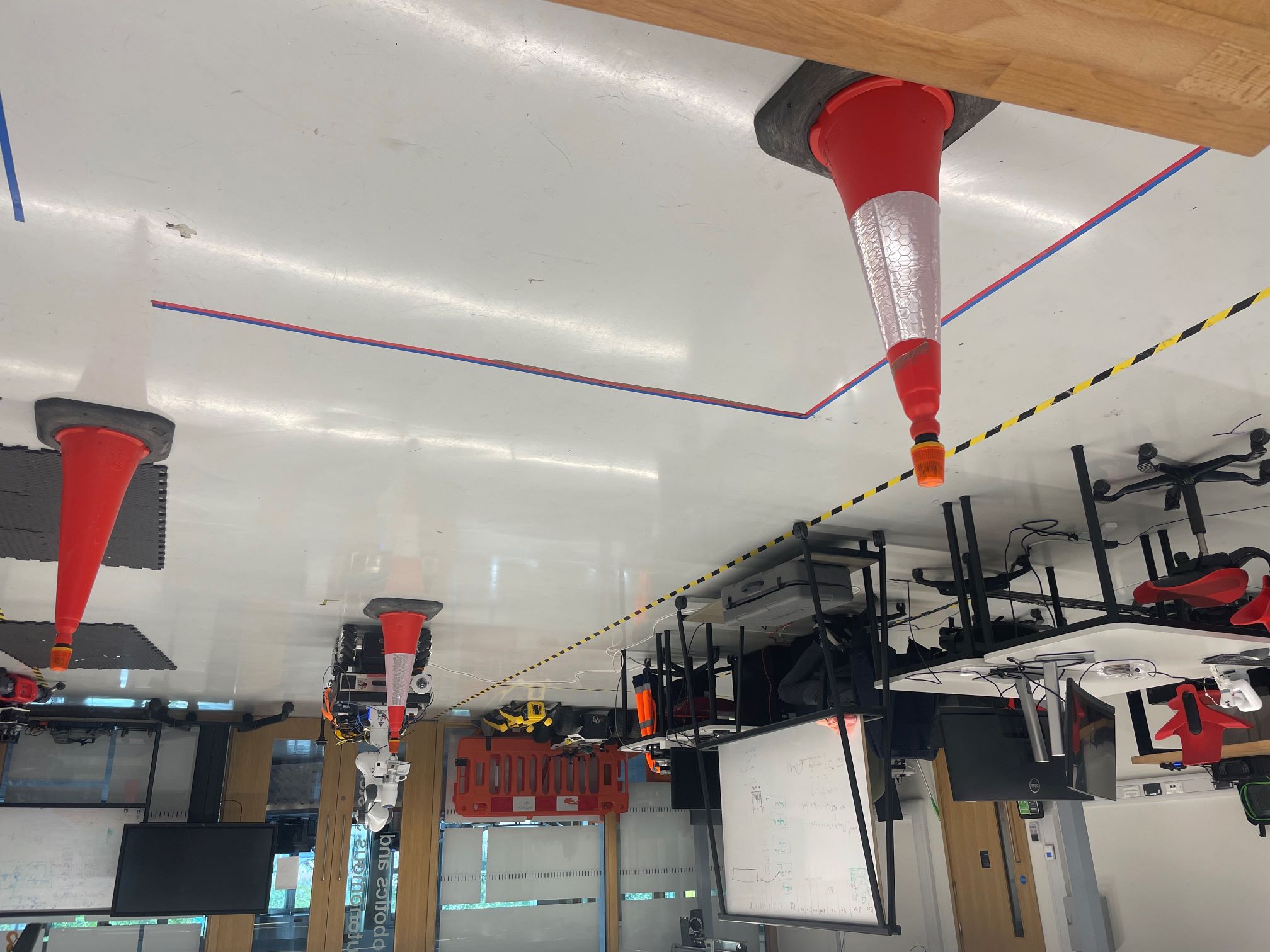} 
    \caption{}
    \label{fig:exp2sub1}
  \end{subfigure}%
  \begin{subfigure}{0.16\textwidth}
    \includegraphics[width=\linewidth , angle=180]{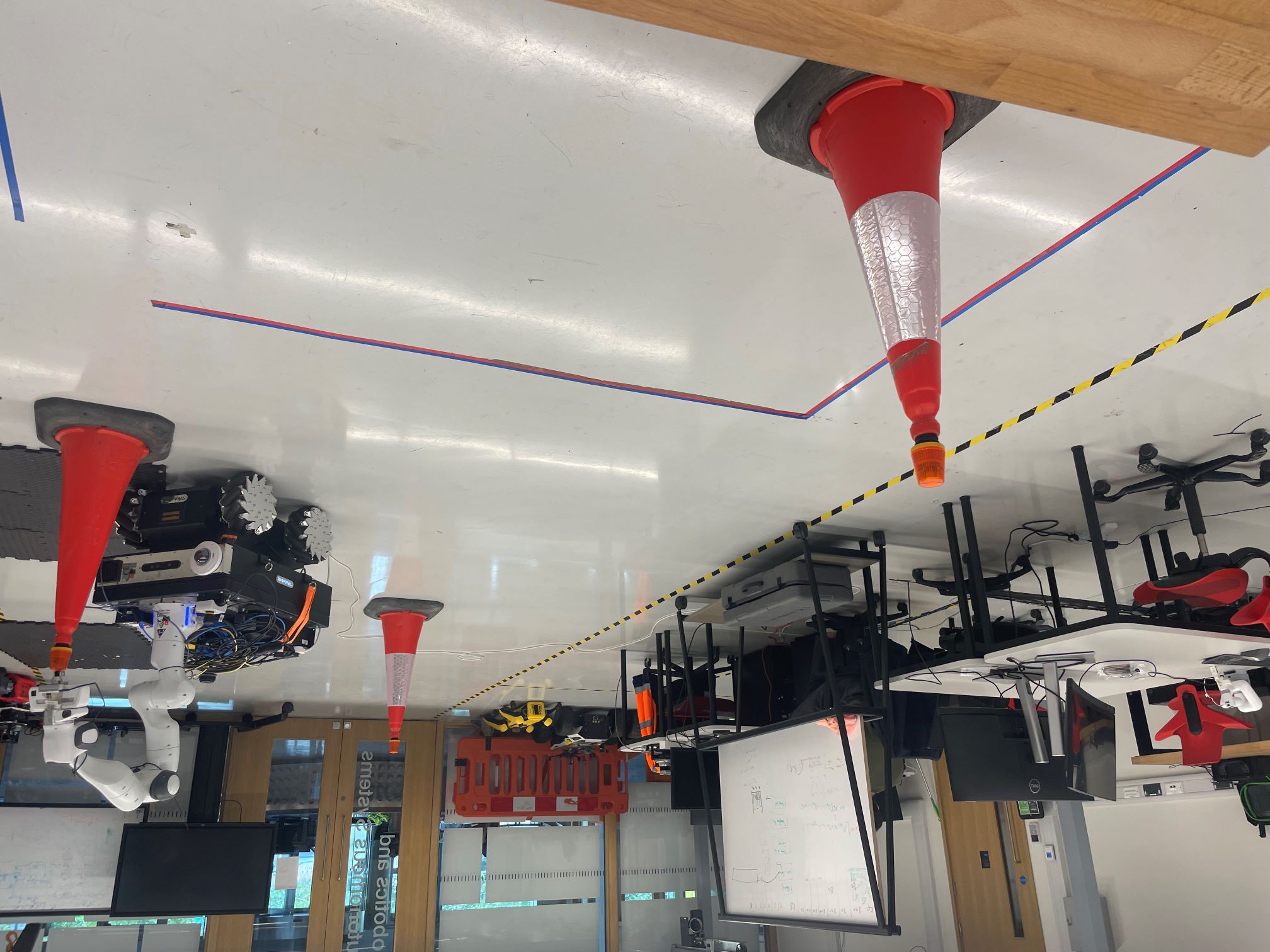} 
    \caption{}
    \label{fig:exp2sub2}
  \end{subfigure}%
  \begin{subfigure}{0.16\textwidth}
    \includegraphics[width=\linewidth , angle=180]{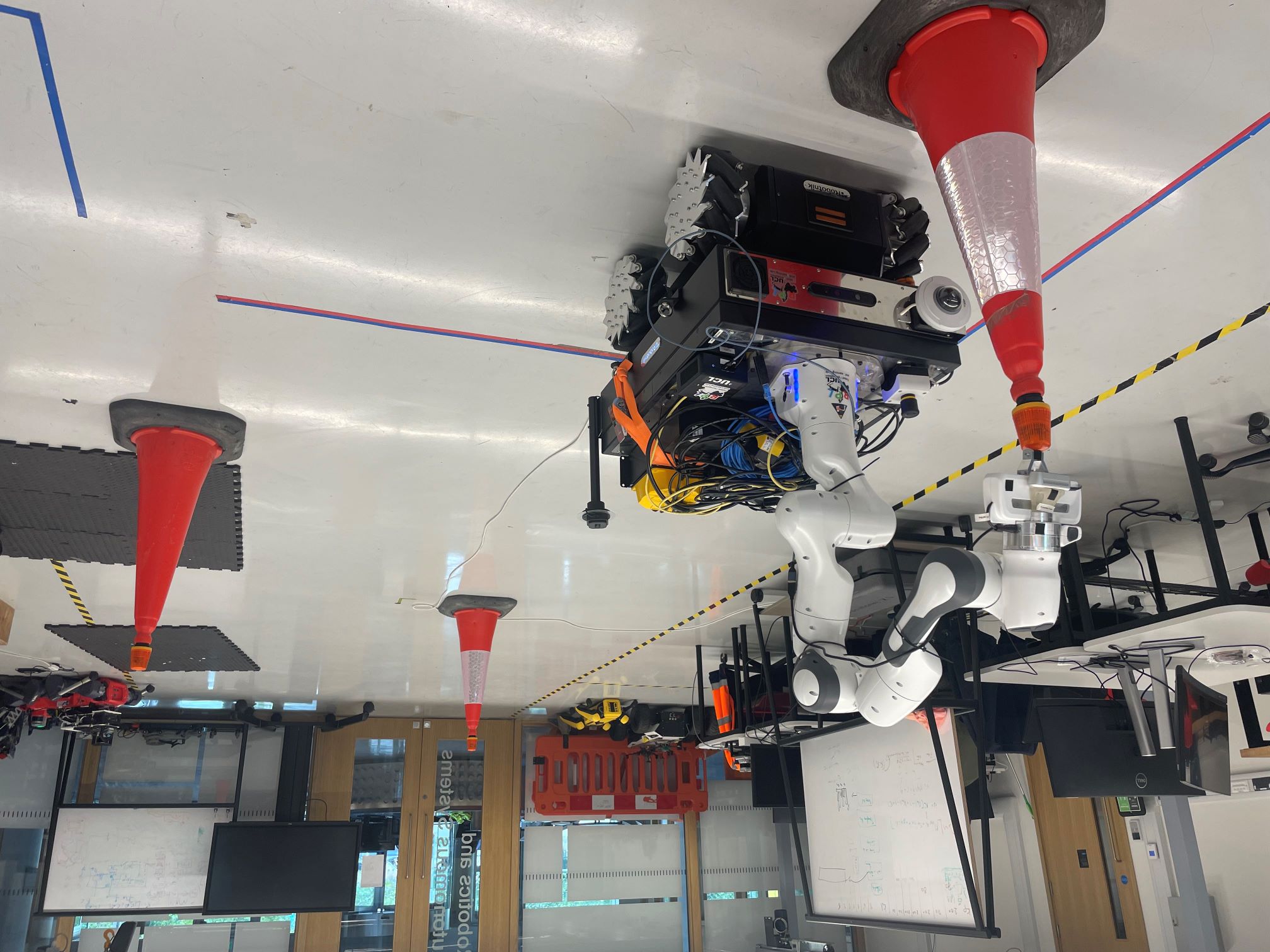} 
    \caption{}
    \label{fig:exp1sub3}
  \end{subfigure}
  \caption{Demonstration of the operator performing two locomanipulation tasks using our teleoperation framework. Panels (a) to (c) illustrate the robot navigating towards a goal destination and successfully reaching a button without any collisions. Panels (d) to (f) depict the operator using the framework to approach each cone, interacting with them to activate the lights placed on top.}
  \label{fig:demo_tasks}
\end{figure}
\subsection{Real Task Design} 

To comprehensively evaluate the capabilities of our experimental setup and obtain qualitative results, we designed two scenarios that required the operator to utilize all aspects of the framework's functionality, as shown in Figure~\ref{fig:demo_tasks}.

In the first scenario, the robot was placed in a controlled environment with various objects arranged on a table, as in Figure~\ref{fig:sctrucut enario1}. After navigating the robot to the table using the locomotion phase, the manipulation phase was tested. The task involved pressing a button on the table, which was obstructed by several household items, hidden from view during the locomotion phase. The operator used the VR interface to teleoperate the manipulator, finding the occluded button and pushing it. These tests assessed the system’s ability to handle occlusions during the manipulation phase, as well as its precision in object manipulation.

In the second scenario, the robot was placed in a controlled environment containing a series of cones, each with a light on top, as in Figure~\ref{fig:demo_tasks} second row. The operator first used the interface to maneuver the robot toward one of the cones, testing the locomotion phase. The manipulation phase was then evaluated by teleoperating the robotic arm to press a button on the cone, which activated the light.

\subsection{Qualitative Results}
The results of the Experimental tests clearly illustrated the strengths of the framework. In the first task, the operator successfully used the locomotion phase to reach the table, and the Gaussian spatter easily captured the occluded button such that in the manipulation phase, it was clearly visible and reachable in the VR environment, as shown in figure \ref{fig:splat example}.

In the second task, the operator easily navigated around the cones during the locomotion phase. Despite the cones being far taller than the table in the first task, the generated Gaussian splat was still of high fidelity and was easily reachable during the manipulation phase.

For both tasks, users easily adapted to the less conventional, third-person view control system for the manipulation phase. They found that the many possible viewpoints provided by the freedom of movement of the interface increased the precision with which they could teleoperate the manipulator. In addition, the generated Gaussian splats were of high enough quality that users could easily distinguish cluttered items such as those in task one. The combination of the high-quality Gaussian splats and flexible control system allowed for the operator to seamlessly perform precise manipulation for both tasks.

\section{CONCLUSIONS}\label{sec:conclusion}







We introduced a VR teleoperation framework for base-manipulator robotic systems that leverages Gaussian splatting to enhance the immersion of the user interface, while maintaining both usability and scalability. Gaussian splatting was found to significantly improve immersion by producing a high-quality 3D model using only an RGB camera mounted on the manipulator, though this came at the expense of increased generation time. The application of Gaussian splatting complemented our proposed teleoperation interface, utilizing a third-person view of the manipulator and the scene, thereby enhancing the immersive experience of the framework, as demonstrated in our user study.

Future work will involve bypassing structure-from-motion to improve the speed of 3D model generation, integrating a dynamic Gaussian splat, and thoroughly testing the framework’s scalability by adapting it for other platforms. 


\addtolength{\textheight}{-12cm}   





\bibliographystyle{IEEEtran}
\bibliography{biblio}

\end{document}